\documentclass[11pt]{article}

\usepackage[a4paper,margin=1in]{geometry}
\usepackage{amsmath,amssymb,amsfonts}
\usepackage{mathtools}
\usepackage{hyperref}
\usepackage{graphicx}
\usepackage{microtype}
\usepackage{enumitem}
\usepackage{booktabs}
\usepackage{titlesec}
\usepackage{xcolor}
\usepackage{url}
\usepackage{caption}
\usepackage{authblk}
\usepackage{tikz}
\usetikzlibrary{arrows.meta,positioning,shapes.geometric,fit,calc}
\usepackage{subcaption}
\usepackage{adjustbox}

\hypersetup{
  colorlinks=true,
  linkcolor=black,
  citecolor=black,
  urlcolor=blue,
  pdfauthor={},
  pdftitle={Background Temperature: Revealing Hidden Randomness in LLMs},
  pdfsubject={LLM determinism, nondeterminism, temperature, batch invariance},
  pdfkeywords={LLMs, nondeterminism, determinism, temperature, batch invariance, softmax}
}

\titleformat{\section}{\large\bfseries}{\thesection}{0.7em}{}
\titleformat{\subsection}{\normalsize\bfseries}{\thesubsection}{0.6em}{}
\title{\textbf{Introducing Background Temperature to Characterise Hidden Randomness in Large Language Models}}
\begin{document}
\author[1]{Alberto Messina}
\author[1]{Stefano Scotta}
\affil[1]{RAI - Radiotelevisione Italiana, Centre for Research, Technological Innovation and Experimentation (CRITS)}
\date{\today}
\maketitle

\begin{abstract}
Even when decoding with temperature $T=0$, large language models (LLMs) can produce divergent outputs for identical inputs. Recent work by Thinking Machines Lab highlights implementation-level sources of nondeterminism, including batch-size variation, kernel non-invariance, and floating-point non-associativity. In this short note we formalize this behavior by introducing the notion of \emph{background temperature} $T_{\mathrm{bg}}$, the effective temperature induced by an implementation-dependent perturbation process observed even when nominal $T=0$. We provide clean definitions, show how $T_{\mathrm{bg}}$ relates to a stochastic perturbation governed by the inference environment $I$, and propose an empirical protocol to estimate $T_{bg}$ via the equivalent temperature $T_n(I)$ of an ideal reference system. We conclude with a set of pilot experiments run on a representative pool from the major LLM providers that demonstrate the idea and outline implications for reproducibility, evaluation, and deployment.
\end{abstract}

\section{Introduction}

A common assumption in LLM deployment is that setting the decoding temperature to $T=0$ (greedy decoding) ensures determinism. However, empirical evidence shows output variability persists under nominally deterministic settings. The recent work in \cite{he2025defeating} argues that nondeterminism in LLM inference often arises from practical systems issues such as varying batch sizes and the lack of batch-invariant kernels, along with floating-point non-associativity and reduction-order effects. This paper proposes a rigorous framing of such effects via the notion of a \emph{background temperature}.

\paragraph{Contributions.} (i) A concise formal model that addresses the phenomenon of nondeterminism as a stochastic effect on the output probability; (ii) a formal definition of background temperature $T_{\mathrm{bg}}$; (iii) the outline of a practical protocol to estimate $T_{\mathrm{bg}}$; (iv) a set of pilot studies illustrating the concept.

\section{Related Work}
\label{sec:related}
The recent work by Thinking Machines Lab provides a systems-first analysis of LLM nondeterminism, emphasizing batch-size variation and batch-invariant kernels for inference; they also explain how floating-point non-associativity and reduction ordering contribute to variability.\,\cite{he2025defeating}. 

In addition to this work, several recent studies have quantified non-determinism in large language model outputs even under settings intended to be deterministic (e.g. temperature \(T=0\), fixed seeds). For example:

\begin{itemize}
  \item \textbf{Atil et al. (2025)} \cite{atil2025non}  systematically evaluate multiple LLMs configured under deterministic settings across zero-shot and few-shot tasks. They observe large accuracy variations (up to ~15\%) across runs with the same input, and show that even the string outputs are often not identical. 

  \item \textbf{Song et al. (2024)} \cite{song2024evaluation} explore how evaluation practices often ignore variability arising from different decoding configurations (greedy vs sampling). They show that even for greedy decoding, evaluation metrics vary, and that alignment methods can help reduce sampling variance. 

  \item \textbf{Ouyang et al. (2023)} \cite{ouyang2023empirical} analyze code generation benchmarks and show that many coding tasks produce different code outputs across repeated prompt invocations, even when using \(T=0\). This confirms that deterministic temperature settings do not guarantee output consistency. 
\end{itemize}

These works align closely with observations from Thinking Machines Lab’s blog \cite{he2025defeating} about system-level implementation factors (batch size, kernel non-invariance, floating point non-associativity, etc.) causing output variation even under nominally deterministic decoding.

While prior work largely documents the existence and magnitude of non-determinism, there remains a gap in formalizing this behavior in terms of an equivalent temperature transformation functional and in proposing standard protocols to measure the effective background randomness. Our work addresses this by introducing the notion of an equivalent temperature \(T_n(I)\) and its expectation \(T_{\mathrm{bg}}\). In the next sections, we transition from formal definitions to a concrete empirical protocol aimed at estimating an \emph{equivalent temperature} $T_n(I)$ induced by implementation noise, and ultimately the background temperature. 

To give a concrete description of what an overall measurement protocol for $T_{\mathrm{bg}}$ would look like, we first describe general criteria for selecting prompts and datasets that are sensitive to small perturbations in model behaviour (including general, task-oriented, and adversarial/synthetic prompts). We then introduce the actual measurement protocol, made up of reference runs under known nonzero temperature settings to calibrate output variability. Following this, based on a suite of quantitative metrics - such as exact-match frequency, first-divergence token index, edit-distance or string similarity, distributional divergence (e.g.\ JS or KL) over next-token / top-k probabilistic outputs, and entropy/confidence measures - we finally outline a fitting procedure to infer $T_n(I)$ by minimizing divergence between outputs under noisy $T=0$ runs and reference nonzero-$T$ runs, and describe how to aggregate over $I$ to compute $T_{\mathrm{bg}}$ with statistical confidence.

\section{Preliminaries and Notation}

Let $D$ denote the token vocabulary with size $|D|$. At generation step $i$, the model produces logits $z \in \mathbb{R}^{|D|}$ and associated probabilities $P(t) \in [0,1]$ that the $i$-th token in the sequence is the $t-th$ token in $D$, such that $\sum_{t = 1}^{|D|}P(t) = 1$ via softmax:

\begin{equation}
  \label{eq:softmax}
  P(t)=P(\tau^t|\tau_{<i}) = \frac{\exp(z_t)}{\sum_{s \in D} \exp(z_s)} \quad \text{for } t = 1, \dots, |D|,
\end{equation}
where $\tau^t$ denotes the $t$-th token in $D$ and $P(\tau^t|\tau_{<i})$ is the probability of generating $\tau^t$ given the sequence of tokens generated up to the $i$-th token.
At $T=0$, the conventional model is greedy decoding by argmax:
\begin{equation}
  \label{eq:greedy}
  \tau^i = \arg\max_{\tau \in D} \, P(\tau \mid \tau_{<i}).
\end{equation}
Decoding at temperature $T>0$ is equivalent to do the same operation but with modified logits $\hat{z} \in \mathbb{R}^{|D|}$:
\begin{equation}
  \label{eq:temp-softmax}
  \hat{P}_T(\tau^i|\tau_{<i}) = \frac{\exp\!\big(\hat{z_i} \big)}{\sum_{s \in D} \exp\!\big(\hat{z_s} \big)}.
\end{equation}

Then the $i$-th token is distributed as some Categorical random variable depending on the probability distribution above, i.e.
\begin{equation}
  \label{eq:categ}
  \tau^i \sim  Categorical (\hat{P}(\tau \mid \tau_{<i})).
\end{equation}
Logits are modified through the randomization effects that are included in the decoding process by the specific LLM implementation. In standard autoregressive language models, the decoding temperature parameter modifies the randomness of next-token selection by adjusting the probability distribution derived from logits. Typically, one scales or transforms the raw (pre-softmax) logits via a temperature parameter and then passes them through softmax to obtain the final distribution for sampling or greedy selection. In general, but as a sufficient assumption for the sake of this work, lower temperatures concentrate probability mass on the most likely tokens, making output more deterministic, while higher temperatures flatten the distribution and increase variability. 

Equivalently, this can be seen as the result of the application of an opportune temperature transformation functional $F_T$:
\begin{equation}
  \label{eq:functional}
  F_T : \mathbb{R}^{|D|} \rightarrow \mathbb{R}^{|D|}, \qquad \hat{P} = F_T(P),
\end{equation}
with the ideal identity limit $F_0(P)=P$. Many implementations use temperature \(T\) so that the model effectively computes something like \(F_T(P)\), a functional transformation of the original token probability vector \(P\), where \(T=0\) corresponds (ideally) to purely greedy decoding, and \(T>0\) allows stochastic sampling.

\section{Modelling Intrinsic Nondeterminism at $T=0$}

As noted by authors in \cite{he2025defeating}, real systems exhibit implementation-dependent perturbations even under $T=0$. Let $I \in \mathcal{I} $ denote the \emph{inference environment} (batch size and composition, concurrency/load, hardware/backends, kernel choices, numeric precision, reduction ordering, etc.) and $F'_T$ the temperature transformation functional of the real system. We model a perturbation $\epsilon_I$, mapping probability distribution over the set $D$ to probability distribution over the same set, that alters the effective distribution as:
\begin{equation}
  \label{eq:perturb}
  F'_0(P)=\epsilon_I(F_0(P)) \approx \epsilon_I(P).
\end{equation}
While $\epsilon_I$ may differ only slightly from $F_0(P)$, in regions where multiple tokens have similar probability mass, even slight changes can flip the argmax in \eqref{eq:greedy} and thus the emitted token sequence.

%\old{We model a perturbation $\epsilon(I)$ that alters the effective distribution as:
%\begin{equation}
%  \label{eq:perturb}
%  F'_0(P)=F_0(P) + \epsilon(I) \approx P + \epsilon(I).
%\end{equation}
%\noindent While $\epsilon(I)$ may be small in norm, in regions where multiple tokens have similar probability mass, even slight changes can flip the argmax and thus the emitted token sequence. Naturally, to achieve a probability distribution as a result of $F'_0$ we assume that $\sum _i \epsilon_i=0$.}

\section{Equivalent Temperature and Background Temperature}

We posit that the perturbation in \eqref{eq:perturb} behaves \emph{as if} decoding were performed by an inference environment - free (ideal) system at a nonzero equivalent temperature $T_n(I)$ :
\begin{equation}
  \label{eq:equiv-T}
  F'_0(P)\approx \epsilon_I(P) \approx F_{T_n(I)}(P).
\end{equation}
%\old{
%\begin{equation}
%  \label{eq:equiv-T}
%  P + \epsilon(I) \approx F_{T_n(I)}(P).
%\end{equation}}
This motivates the following definition.

\paragraph{Definition (Background temperature).}
The \emph{background temperature} of an LLM implementation is the expected equivalent temperature induced by the inference environment under nominal $T=0$:
\begin{equation}
  \label{eq:Tbg}
  T_{\mathrm{bg}} \triangleq \mathbb{E}_{I \in \mathcal{I}}\!\left[ T_n(I) \right].
\end{equation}

Intuitively, $T_{\mathrm{bg}}$ captures the \emph{implicit} randomness in a deployment stack even when the user selects $T=0$.

\section{Estimating $T_n(I)$ and $T_{\mathrm{bg}}$ Empirically}
\label{sec:estimation}
The problem with the definition given in  \eqref{eq:Tbg} is that the inference environment-free (ideal) system may be not at hand. In fact, the key challenge in estimating $T_n(I)$ is that it  requires comparing to a perfect, deterministic reference - which may not exist in practice. To make $T_n(I)$ calibration feasible without an unattainable ideal, one can first identify a quasi-ideal environment: for example, by using inference pipelines with batch-invariant kernels (in normalization, matrix multiplication, attention), fixed numeric precision, minimal or single-request concurrency, and deterministic configuration flags. Thinking Machines Lab demonstrates that replacing standard kernels with batch-invariant ones drastically reduces output divergence under zero temperature \cite{he2025defeating} Similarly, \cite{shanmugavelu2024fpna} show that floating-point non-associativity and asynchronous parallel reductions are major sources of run-to-run variability, and that enforcing deterministic alternatives significantly stabilizes inference and scientific computing pipelines. Based on this evidence, one can anchor measurement of $T_n(I)$ relative to such quasi-ideal baselines, or employ multiple such baselines (differing in hardware, kernel implementation, or precision) to absorb uncertainty. Further, measuring various output statistical distributions (rather than only output strings) allows matching of environments $I$ to baselines via statistical divergence metrics, reducing sensitivity to rare argmax flips. Reporting $T_n$ together with such baseline variances yields operationally meaningful estimates even in the absence of a perfect oracle.  

Another practical way to assess the background temperature of an online model (e.g. ChatGPT) is to use a local installation of another model (e.g. Llama) as a benchmark reference. The local model must be configured to be as deterministic and stable as possible—fixed precision, consistent batch sizes, kernel implementations that do not alter behavior when batch shape changes, deterministic reduction orders, disabled non-deterministic/autotuned operations. This reference becomes a baseline environment that approximates ``ideal behavior''. Then, by comparing output distributions from the online model versus those from the stable local model, one can compute how far the online model’s behavior diverges, for example via measures like Jensen-Shannon divergence or KL divergence. By finding what temperature setting of the local model would make its distribution match the diverged distribution of the online model, it is possible to infer an equivalent temperature for the online model in that environment. Repeated across many prompts and local configurations, this yields an estimate of the online model’s background temperature, together with uncertainty bounds. This method avoids relying on an unattainable perfect system, by using the best stable reference you can build.

\begin{figure}[ht]
    \centering
    \includegraphics[width=1.0\linewidth]{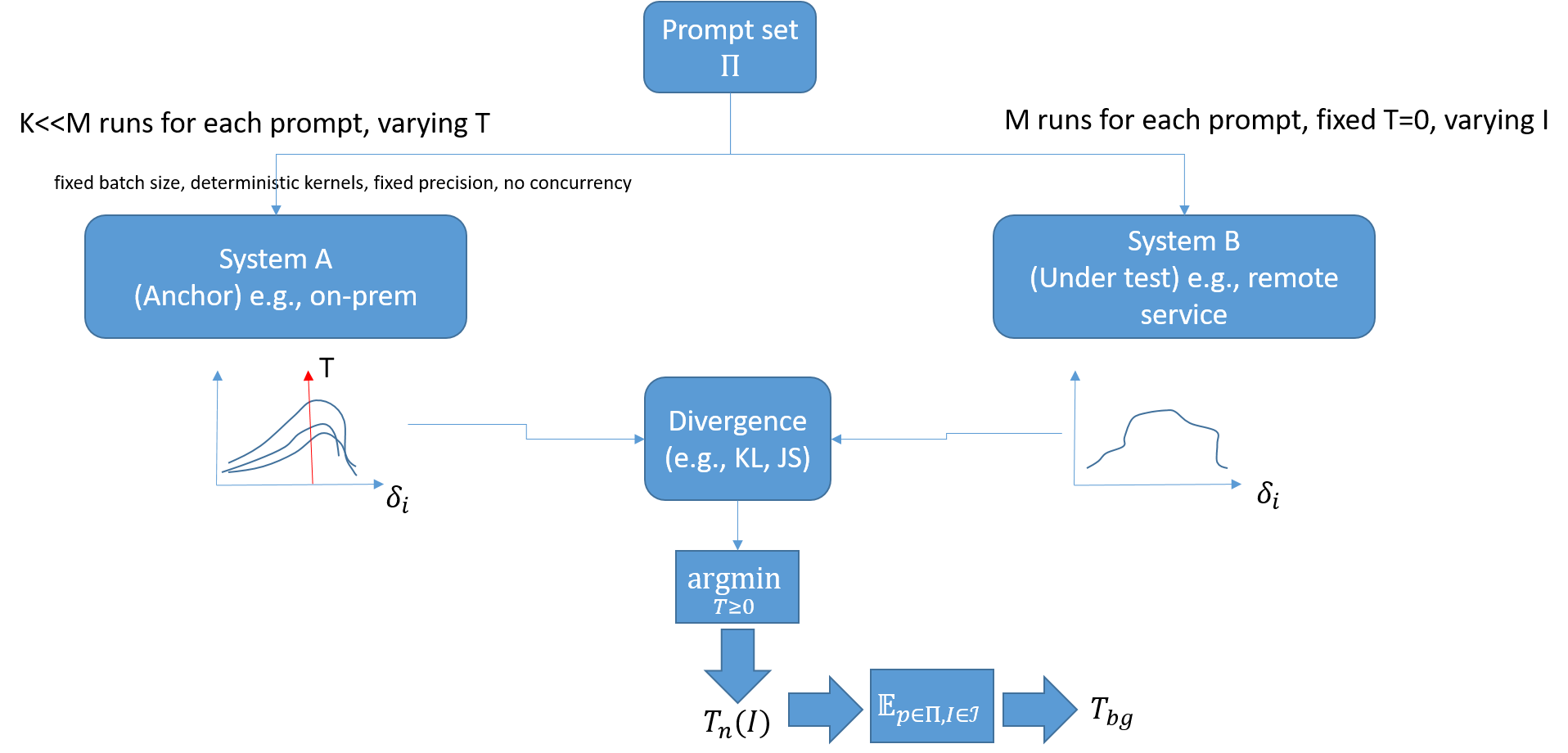}
    \caption{\small Measuring protocol.}
    \label{fig:flow}
\end{figure}

With these considerations in mind, we can outline a practical protocol to estimate $T_n(I)$ and $T_{\mathrm{bg}}$ which is pictorially described in Figure \ref{fig:flow}.

\subsection{Prompt Sets and Datasets}
The first element of the protocol is constituted by a relevant prompt set $\Pi$, an element of the theoretical set of all the possible combinations of prompts $\mathcal{P}$. To explore the full range of behavior of the system under test, the suggestion is to use a diverse evaluation suite, e.g.:
\begin{itemize}[leftmargin=1.5em]
  \item \textbf{General generation} prompts (short/long, common/rare vocab).
  \item \textbf{Task benchmarks}: QA (e.g., SQuAD\cite{price2025document}/TriviaQA\cite{joshi-etal-2017-triviaqa}), summarization, close, and short-format classification. Code-generation prompts if applicable.
  \item \textbf{Edge/adversarial} prompts: long contexts, rare tokens, near-ties among top-$k$ token probabilities.
  \item \textbf{Synthetic} prompts engineered to create finely balanced next-token choices.
\end{itemize}

\subsection{Controlling the Inference Environment $I$}

Run repeated inference (e.g., $M\ge 50$ per prompt) at $T=0$ while varying $I$ along axes known to influence nondeterminism:
\begin{itemize}[leftmargin=1.5em]
  \item \textbf{Batch structure}: batch size, e.g. $\in \{1,2,4,8,16,32,\dots\}$; co-batching with other prompts vs.\ serial.
  \item \textbf{Concurrency/load}: single request vs.\ many simultaneous requests.
  \item \textbf{Hardware/backends}: GPU types, CPU vs.\ GPU, precision (fp16/bf16/fp32), kernel implementations (batch-invariant vs.\ standard).
  \item \textbf{Numerics}: reduction order, deterministic flags in frameworks, fused vs.\ unfused kernels.
\end{itemize}
For remote systems, for which it may be impossible or impractical to govern the inference environment, one can assume that prolonged and repeated operation is a good way to sample the inference environment statistical distribution.
\subsection{Reference Runs at Known Temperatures}

Under a \emph{stable} environment $I_{stable}$, e.g. a local anchor system, run the same prompts e.g. at a grid of $T  \in \{0,\ldots, 1, \ldots\}$ to build a mapping between $T$ and output-variability statistics. As noted earlier, this stable environment can either be a specific configuration of the system under test or another anchor used as reference. Given that the anchor configuration is supposed to be stable for what concerns the inference environment, a lower number $K$ of runs for each prompt in the prompt set should suffice.

\subsection{Variability Metrics}\label{varmetr}
The key element of the protocol is the set of metrics used to obtain the association between the sought-for background temperature parameter for the system under test and the reference measurements on the anchor system. Since th $T_{\mathrm{bg}}$ is thought as a genereic high level account of the system's nondeterminism, metrics should be content-agnostic. Furthermore, since different systems are trained independently, it is practically certain that the same prompt would produce different outputs even under strict deterministic configurations.
For example, for each prompt, and across the $M$ (or $K$) runs of Figure \ref{fig:flow}, compute process parameters like e.g.:
\begin{itemize}[leftmargin=1.5em]
  \item \textbf{Exact-match rate}: fraction of runs producing identical strings for the same prompt.
  \item \textbf{First-divergence index}: position of first token mismatch across pairs of runs.
  \item \textbf{Edit distance} first order and second order statistics between different outputs.
  \item \textbf{Distributional divergence}: e.g., symmetrized KL or JS divergence between empirical next-token distributions (top-$k$) across runs.
  \item \textbf{Entropy} of next-token distributions.
\end{itemize}
Then, for each variability metric computed across the runs, construct a multidimensional distribution $f$ that captures the values of the variability metrics for the system considered. In particular, we'll denote by $f_T(I_\text{stable})$ and $g(I)$ respectively the distribution of the variability metrics for the reference system when the temperature is $T$ and for the system under test set at temperature $0$. Note that these distributions depend on multiple factors, including the specific LLMs used; for notational simplicity, we omit these dependencies.

%\old{
%Then, build a multivariate distribution $\Delta(\delta_1,\ldots,\delta_M)$ where $\delta_i$ is the $i-th$ process parameter. Each distribution $\delta_i$ thus represents the variability across $\Pi$ of the $i-th$ process parameter.}

\subsection{Estimators for $T_n$ and $T_{\mathrm{bg}}$}\label{estimations}
As explained in Section \ref{sec:estimation}
the ideal reference system does not exist. However, it is possible to estimate $T_n$ using some reference model running in an environment $I_{stable}$ as stable as possible. In particular, for a reference LLM $\ell$, it is possible to compute an estimator $\hat T_n^{\ell} = \hat T_n^{\ell} (I, \Pi)$ of $T_n$, for each $I$ in the set of environments considered $\mathcal{\tilde I} \subseteq \mathcal{I}$ and each $\Pi$ in the set of all the collections of prompts considered $ \mathcal{\tilde P}\subseteq \mathcal{P}$, as 
\begin{equation}
  \label{eq:fit}
  \hat T_n^{\ell} = \arg\min_{T\geq 0} \; \mathcal{D} \Big(f_T(I_\text{stable}), g(I)\Big),
\end{equation}
where $\mathcal{D}$ is a chosen divergence (e.g., JS or KL divergence, or a weighted combination) applied to the variability distributions
 $g(I)$ and $f_T(I_\text{stable})$, corresponding respectively to the system under test and to the reference system based on $\ell$ (see Section \ref{varmetr}). Therefore, it is possible to compute an estimate $\hat T_{\mathrm{bg}} = \hat T_{\mathrm{bg}}(\ell)$ of $T_{\mathrm{bg}}$, for each $\hat T_n$, as 
\begin{equation}
    \hat T_{\mathrm{bg}}(\ell) =\frac{1}{|\mathcal{\tilde I}|}\frac{1}{|\mathcal{\tilde P}|} \sum_{I \in \mathcal{\tilde I}} \sum_{\Pi \in \mathcal{\tilde P}} \hat T_n^{\ell} (I, \Pi),
\end{equation}
where $|\mathcal{\tilde I}|$ and $|\mathcal{\tilde P}|$ denote, respectively, the number of all the $I$ and $\Pi$ considered. To further improve robustness, we repeat the same process across a set of different reference LLMs $\mathcal{L}$ and take the average\footnote{Beyond the average estimate, the availability of multiple reference LLMs and configurations also allows the computation of higher-order moments and confidence intervals, providing a more precise characterization of the uncertainty associated with this kind of estimate.} 
\begin{equation}\label{overline}
    \overline T_{\mathrm{bg}} = \frac{1}{|\mathcal{L}|}\sum_{\ell\in\mathcal{L}}\hat T_{\mathrm{bg}}(\ell),
\end{equation}
where $|\mathcal{L}|$ denotes the number of different $\ell$ (LLMs) used.
Theoretically, as the set of reference LLMs $\mathcal{L}$, prompts, environments, and variability metrics grows, we can expect $\overline T_{\mathrm{bg}}$ to converge to the true $T_{\mathrm{bg}}$, as defined in \eqref{eq:Tbg}.
%\old{
%\subsection{Fitting $T_n(I)$ and Aggregating $T_{\mathrm{bg}}$}
%For each environment $I$, fit
%\begin{equation}
%  \label{eq:fit}
 % T_n(I) \;=\; \arg\min_{T\geq 0} \; \mathcal{D}\!\Big(\mathrm{\Delta}(T{=}0, I)\;,\; \mathrm{\Delta}(T, I_{\text{stable}})\Big),
%\end{equation}
%where $\mathcal{D}$ is a chosen divergence on variability metrics (e.g., JS or KL divergence, or a weighted combination of metrics), and $I_{\text{stable}}$ is a fixed environment used to calibrate $T$. Finally, report
%\begin{equation}
 % T_{\mathrm{bg}} \;=\; \mathbb{E}_{I,\Pi}[T_n(I)], \qquad
%  \mathrm{Var}_{I,\Pi}[T_n(I)], \qquad \text{and confidence intervals.}
%\end{equation}
%Notice that in an ideal deterministic system $T_n(I)$ would be always $0$ since the process parameter distribution would be independent from environment $I$ at any value of $T$.}

\subsection{Engineering to Reduce $T_{\mathrm{bg}}$}
Once for a certain system the background temperature $T_{\mathrm{bg}}$ is available, several mechanisms can be put in place to mitigate its effect. For example, empirical and systems work suggests several interventions:
\begin{itemize}[leftmargin=1.5em]
  \item \textbf{Batch-invariant kernels} for core ops (matmul, attention, RMSNorm) to prevent batch-shape–dependent numerics \cite{he2025defeating}.
  \item \textbf{Deterministic reductions} and stable accumulation orders where feasible \cite{shanmugavelu2024fpna}.
  \item \textbf{Consistent pipelines}: fix kernel configs across shapes; avoid opportunistic algorithm switching that alters reduction paths \cite{ravi2025reproducibility}.
  \item \textbf{Deterministic flags} in frameworks and careful precision selection \cite{shanmugavelu2024fpna}.
  \item \textbf{Operational controls}: cap concurrency or shape buckets to reduce co-batching variability \cite{he2025defeating}.
\end{itemize}
Ablation studies can further determine what intervention is impacting the most on the background temperature. This transforms the outlined protocol into an iterative practice aimed at controlling the nondeterministic characteristics of the system in use, as opposed to a mere observation of an empirical phenomenon.

\section{Pilot Experiments}
In this section, we present some experiments to validate the theory presented in this work. In particular, in Section \ref{firstexp}, we present a simple pipeline for estimating the background temperature for a given model. After that, we present additional experiments that could clarify and add elements to analyze the background temperature.

\subsection{Basic pipeline for estimating $T_{\mathrm{bg}}$}\label{firstexp}
Here we perform a pilot experiment to estimate $T_{\mathrm{bg}}$ for the OpenAI model \textit{gpt-4.1-nano} accessed via the Microsoft Azure AI services with temperature $T=0$ (i.e., considering it as System B in Figure \ref{fig:flow}). Note that, being the model used a via third part service, we can not control the inference environment $I$ but only the temperature. 
The prompt set $\Pi$ used is composed of the first 200 questions of the dataset \textit{truthful\_qa} \footnote{\url{https://huggingface.co/datasets/truthfulqa/truthful_qa}} (see \cite{truthfulqa}). 

The reference LLM $\ell$, playing the role of System A in Figure \ref{fig:flow}, is Hugging-Face LLM \textit{SmolLM3-3B}\footnote{\url{https://huggingface.co/HuggingFaceTB/SmolLM3-3B}} (see \cite{smollm3}). As outlined in previous sections, we selected representative temperature values $\Theta$  sampled in increments of $0.01$ from $0$ to $0.2$, in increments of $0.05$ from $0.2$ to $0.5$ and in increments of $0.1$ from $0.5$ to $1$, i.e. $$\Theta = \{0,0.01,\dots, 0.19, 0.2, 0.25, \dots, 0.45, 0.5, 0.6, \dots, 0.9, 1\}.$$ For each $T\in\Theta$, we generated $K=32$ responses, limited to $32$ tokens, with the reference LLM for each of the $200$ prompts in $\Pi$. As variability metric (see Section \ref{varmetr}), we used the exact-match fraction, i.e. for each temperature considered and each prompt in $\Pi$, we computed the maximum fraction of identical answers among the $32$ generated. In this way, for each $T \in \Theta$ we obtained $200$ values in the interval $[1/32, 1]$, which constitute the discrete distribution $f_{T}$ of the exact-match fraction for that temperature in the answers given by the reference LLM.

\begin{figure}[!htbp]
    \centering
    % prima riga: tre figure
    \begin{subfigure}[b]{0.32\textwidth}
        \centering
        \includegraphics[width=\textwidth]{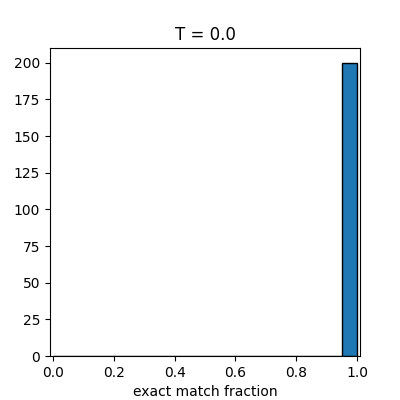}
        \label{fig:sub1}
    \end{subfigure}
    \hfill
    \begin{subfigure}[b]{0.32\textwidth}
        \centering
        \includegraphics[width=\textwidth]{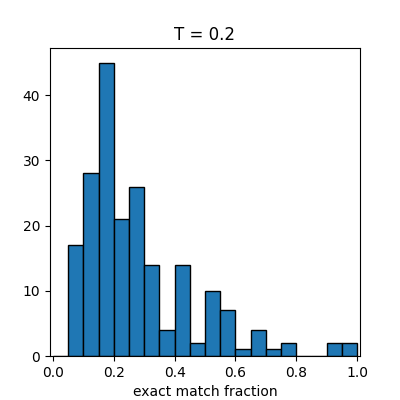}
        \label{fig:sub2}
    \end{subfigure}
    \hfill
    \begin{subfigure}[b]{0.32\textwidth}
        \centering
        \includegraphics[width=\textwidth]{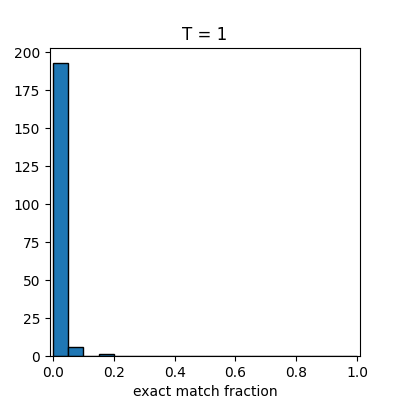}
        \label{fig:sub3}
    \end{subfigure}

    % seconda riga: una figura larga
    \begin{subfigure}[b]{\textwidth}
        \centering
        \includegraphics[width=\textwidth]{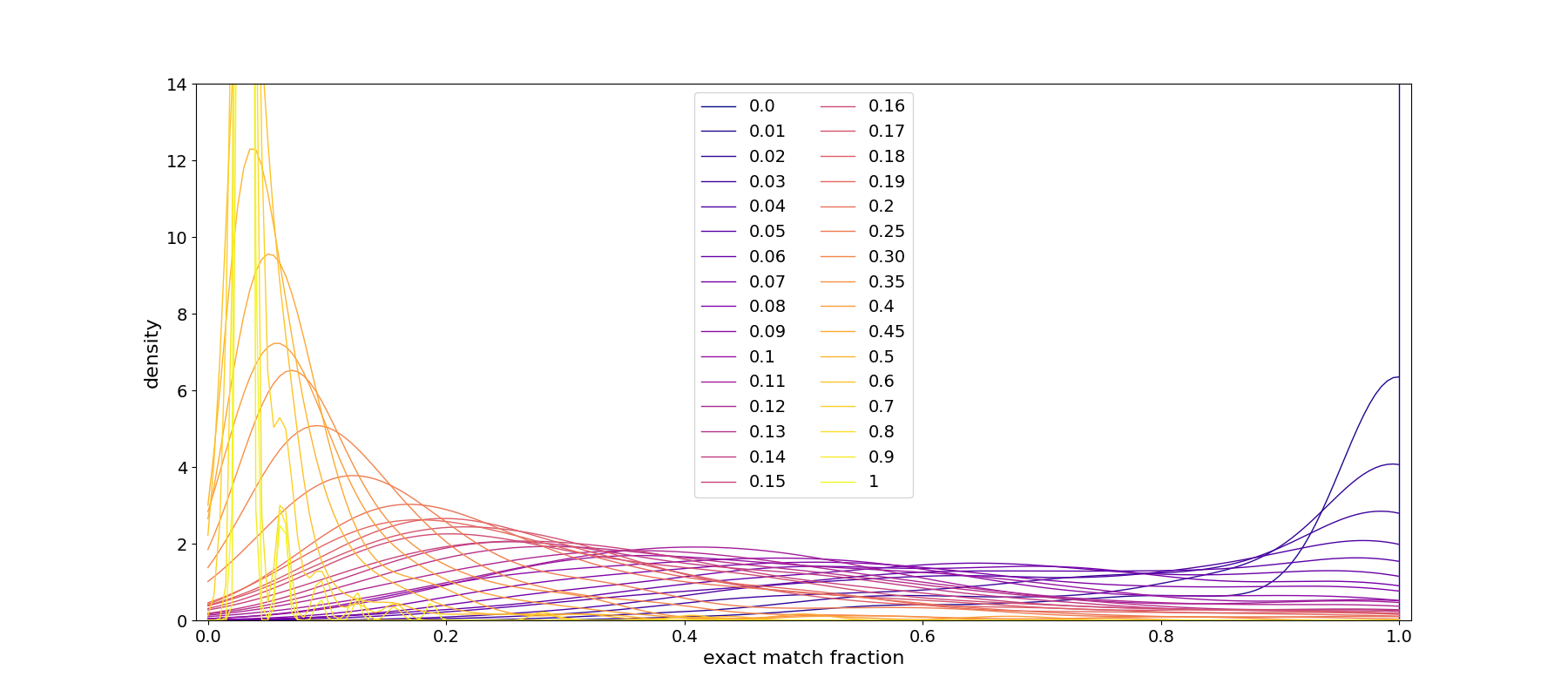}
        \label{fig:sub4}
    \end{subfigure}

    \caption{\small Distribution of exact-match fractions obtained from the reference LLM answers. 
Top row (from left to right): histograms representing the distributions $f_{0}$, $f_{0.2}$ and $f_1$.  
Bottom row: kernel density estimates of the exact-match fraction for all sampled temperatures in 
$\Theta$. Note that for $T=0$, the density is represented as a vertical line because all answers are identical, so the density is entirely concentrated at 1, forming a Dirac delta.}
    \label{fig:discrete_distribution}
\end{figure}
In Figure \ref{fig:discrete_distribution}, these distributions are graphically represented, showing how the density estimate shifts from a delta concentrated at $1$ when the temperature is $0$ - indicating that all answers are identical - to a distribution with most of its mass near $0$, indicating that the answers tend to be unique.

\begin{figure}[ht]
    \centering
    \includegraphics[width=1.0\linewidth]{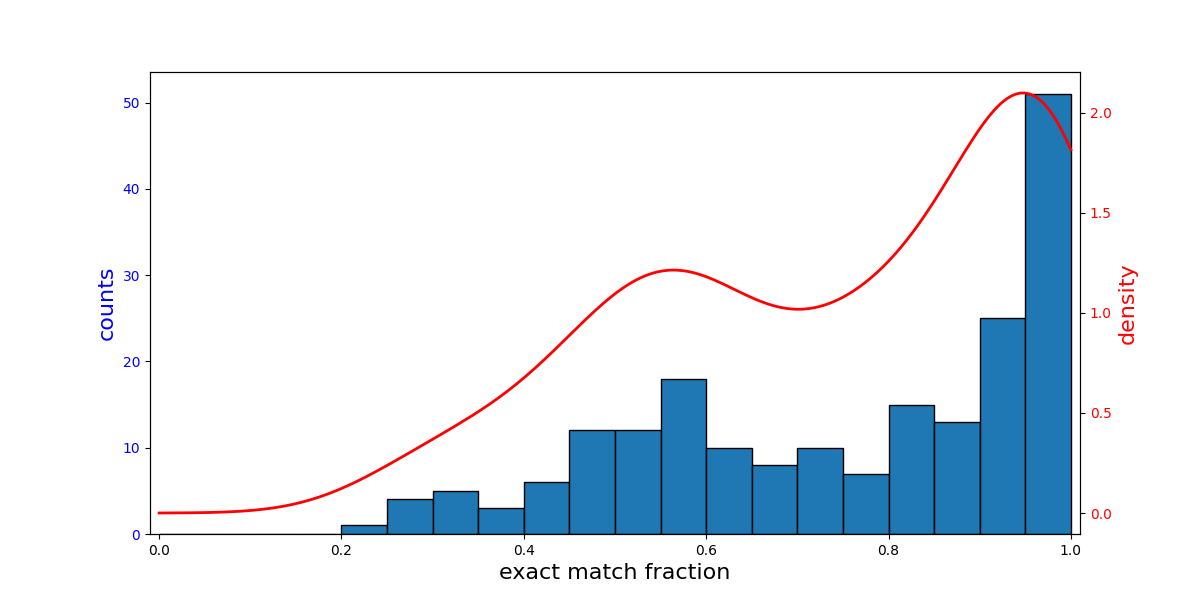}
    \caption{\small Discrete distribution $g$ of the fraction of identical answers given by the LLM under test, \textit{gpt-4.1-nano}, to the prompts in $\Pi$. The distribution is shown both as histograms (with the $y$-axis on the left) and as a kernel density estimate (with the $y$-axis on the right).}
    \label{fig:gpt_distr}
\end{figure}

After computing the reference distributions $f_T$ for $T \in \Theta$ of the chosen variability measure, we computed the same for the model for which we want to estimate $T_{\mathrm{bg}}$, i.e., \textit{gpt-4.1-nano}, accessed via the Microsoft Azure AI services. To do this, we prompted the model $100$ times for each of the $200$ prompts in $\Pi$, but this time setting the temperature at $T=0$ and limiting the answers to $32$ tokens, as done for the reference system. Then, analogously to the procedure for the reference system, for each prompt in $\Pi$ we computed the maximum fraction of identical answers provided by \textit{gpt-4.1-nano}. These $200$ values, in $[1/100, 1]$, form the discrete distribution $g$ (see Figure \ref{fig:gpt_distr}) that we need to compare with the reference distributions computed in system A (see \eqref{eq:fit}).

\begin{figure}[h!]
    \centering
    % --- FIGURA KSsmollM ---
    \begin{subfigure}{0.69\textwidth}
        \centering
        \includegraphics[width=\linewidth]{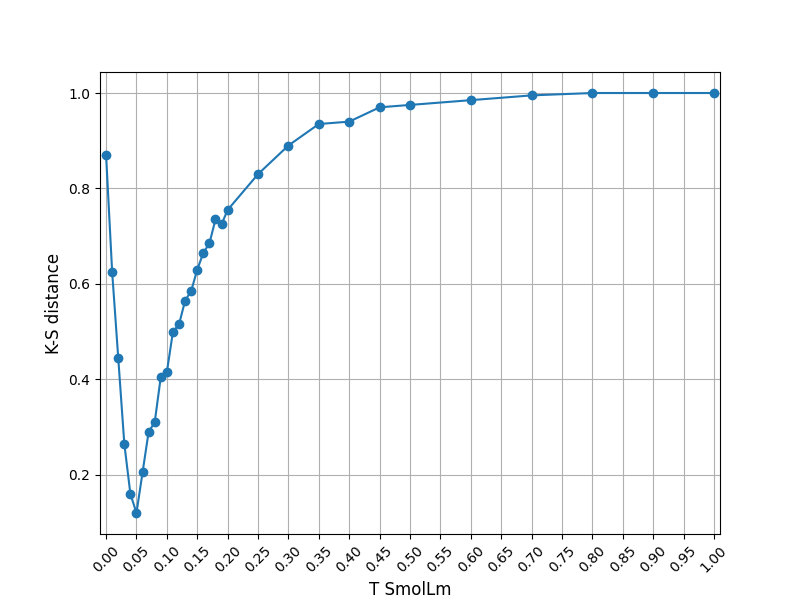} % <-- nome del file della figura
        \caption{K-S distance}
        \label{fig:KSsmollM}
    \end{subfigure}
    \hfill
    % --- TABELLA ---
    \begin{subfigure}{0.28\textwidth}
    \centering
    \begin{tabular}{c c}
    \hline
    $T$ & K-S \\
    \hline
    0.00 & 0.870  \\
    0.01 & 0.625 \\
    0.02 & 0.445 \\
     0.03 & 0.265 \\
    0.04 & 0.160  \\
    \textbf{0.05} & \textbf{0.120} \\
     0.06 & 0.205 \\
     0.07 & 0.290 \\
     0.08 & 0.310 \\
     0.10 & 0.415 \\
     0.15 & 0.630 \\
     0.30 & 0.890 \\
     0.50 &0.975 \\
     1.00 &1.000 \\
    \hline
    \end{tabular}
       \caption{Sample of the K-S values}
        \label{tab:values}
    \end{subfigure}

    \caption{\small {K-S distances between $f_T$ and $g$ at different $T \in \Theta$ for the tested model \textit{gpt-4.1-nano}. Plot of all the values (a). Exact values for a sample of temperatures (b). The tested model background temperature estimate is $0.05$.}}
    \label{fig:KSsmollM_combined}
\end{figure}

In order to compare the discrete distributions of observations, $f_T$ for $T \in \Theta$ and $g$, we chose to use the Kolmogorov–Smirnov (K-S) distance, which is equal to $0$ for identical distributions and $1$ for completely different ones. The computed values of K-S distance are reported in Figure \ref{fig:KSsmollM_combined}.

\begin{figure}[!ht]
    \centering
    \includegraphics[width=1.0\linewidth]{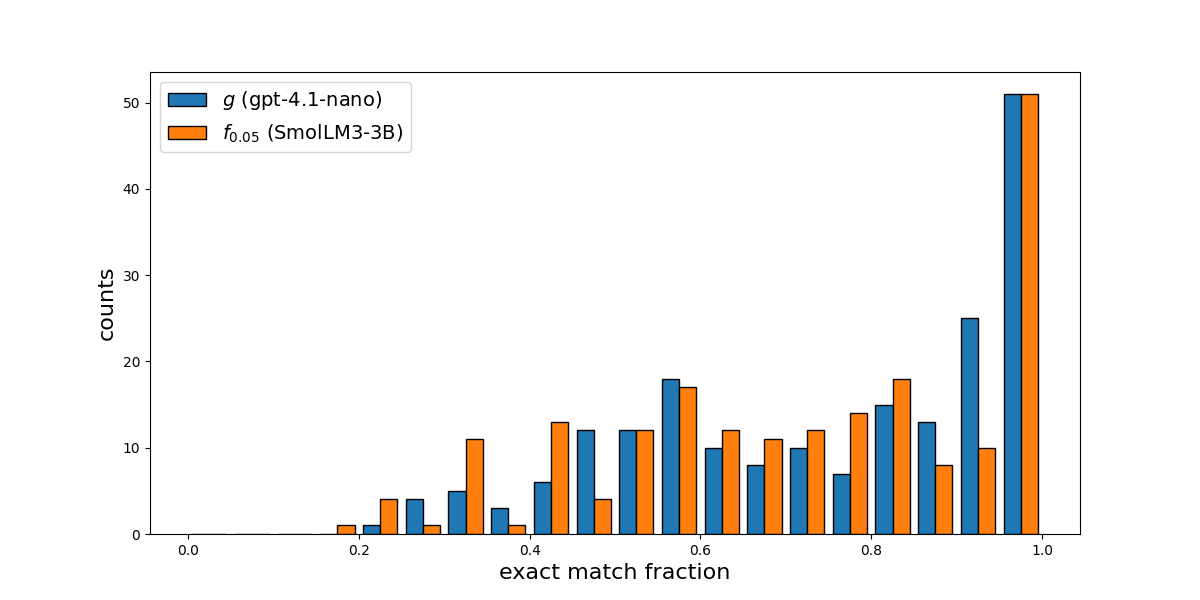}
    \caption{\small Side-by-side histograms of the two empirical discrete distributions, $g$ and $f_{0.05}$, which is the closest to $g$ among $\{f_{T} : T \in \Theta\}$ in terms of K-S distance.}
    \label{fig:2ist}
\end{figure}
From the values in Table \ref{fig:KSsmollM_combined} (b), we can conclude that the estimator of $T_{\mathrm{bg}}$ found in this experiment is $\hat T_{\mathrm{bg}}(\ell)=0.05$ (which, in this simple case, coincides with $\overline T_{\mathrm{bg}}$), as this is the case where $f_T$ is closest to $g$, considering only the reference distributions computed from $T \in \Theta$. Figure \ref{fig:2ist} shows the two matching histograms. Ideally, this experiment should be repeated using a wider range of $T$ values - especially lower ones - more prompts, fewer token limits, and different variability metrics (see Sections \ref{varmetr} and \ref{estimations}). However, the purpose of this pilot experiment was simply to demonstrate the full procedure to estimate $T_{\mathrm{bg}}$.

\subsection{Extending the reference model set $\mathcal{L}$}\label{extendingL}
One of the possibilities for making the estimate of the $T_{\mathrm{bg}}$ more robust is to add reference models, i.e. extend the set $\mathcal{L}$ introduced in Section \ref{estimations}. In particular, we used the LLM \textit{Llama-3.2-3B-Instruct}\footnote{\url{https://huggingface.co/meta-llama/Llama-3.2-3B-Instruct}} and made it answer 32 times to the same 200 prompts (the same set $\Pi$ used in Section \ref{firstexp}), limiting the answers to 32 tokens, analogously to what we did for \textit{SmolLm3-3B} in Section \ref{firstexp}, for each $T \in \tilde\Theta = \Theta \cup\{1.05,1.1, \dots, 1.45, 1.5\}$. So, hereinafter the set of reference LLM $\mathcal L = \{\text{smoll}, \text{llama}\}$, where $\text{smoll}$ and $\text{llama}$ stand, respectively, for \textit{SmolLm3-3B} and \textit{Llama-3.2-3B-Instruct}. Then we computed for $\text{llama}$ the analogous to $f_T$ for $\text{smoll}$, i.e. the reference discrete distribution $\tilde f_T$ for all values of $T \in \tilde \Theta$ of the maximum fraction of identical answers among the 32 generated for each prompt in $\Pi$ by $\text{llama}$ with temperature $T$. 

Then it is possible to compute the value of $\hat T_{\mathrm{bg}}(\text{llama})$ for the model under test (\textit{gpt-4.1-nano}), exactly as we did for $\hat T_{\mathrm{bg}}(\text{smoll})$ in Section \ref{firstexp}. In particular, the minimum value of the K-S distance between $g$ and $\tilde f_T$, for $T \in \tilde \Theta$ is $0.155$ and it is reached when $T=0.1$, so $\hat T_{\mathrm{bg}}(\text{llama}) = 0.1$.
Hence, we can compute a more robust value of $\overline T_{bg}$ for the LLM under test, using \eqref{overline}:
\begin{equation}
    \overline T_{bg} = \frac{\hat T_{\mathrm{bg}}(\text{smoll}) + \hat T_{\mathrm{bg}}(\text{llama})}{2} = 0.075.
\end{equation}

\begin{figure}[!ht]
    \centering
    \includegraphics[width=1.0\linewidth]{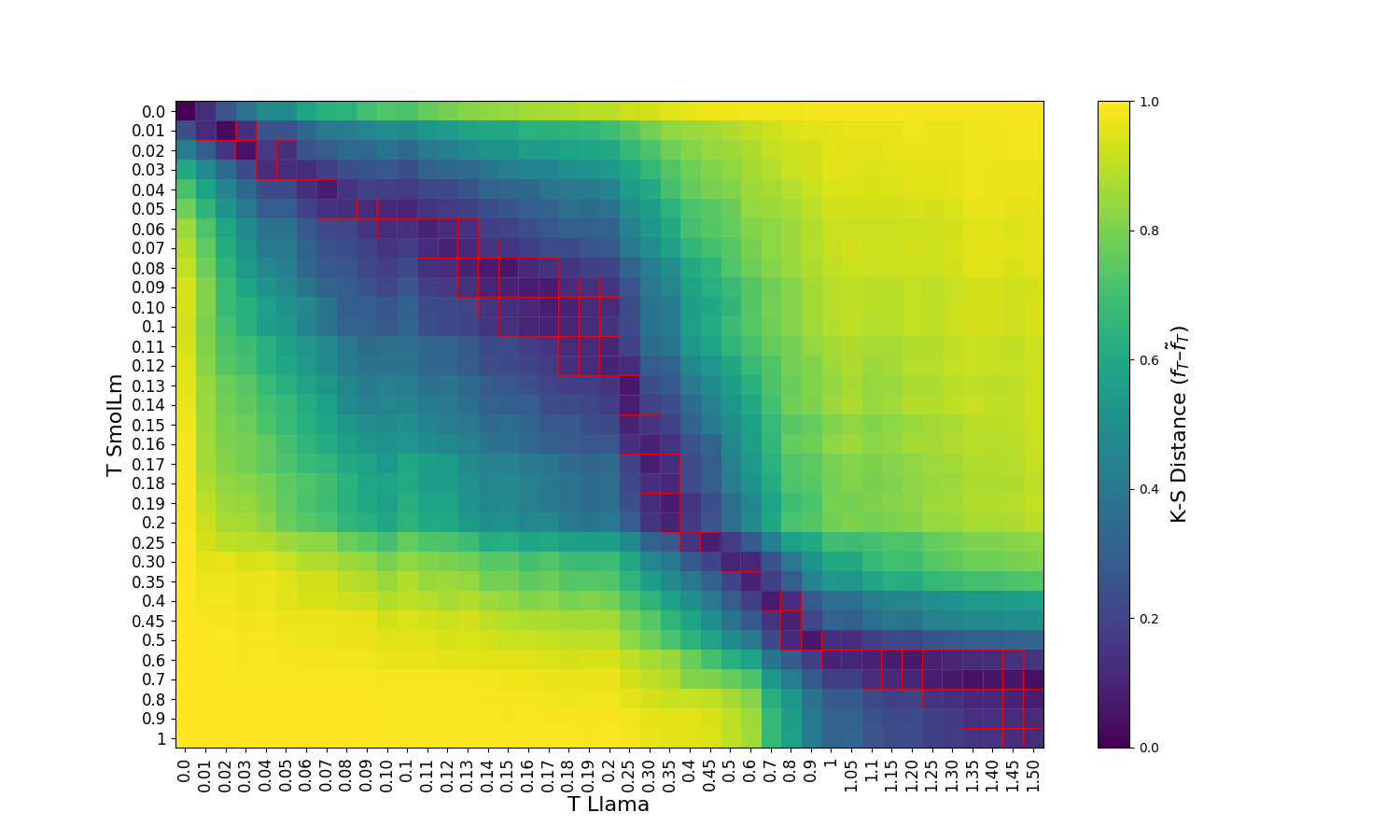}
    \caption{\small Heatmap representing the K-S distance between $f_T$ and $\tilde f_{T'}$ for $T\in\Theta$ and $T' \in \tilde \Theta$. In particular the cell on the $t$-row and $t'$-column represent the K-S distance between $f_t$ and $\tilde f_{t'}$. Cells with red borders correspond to K-S distance values less or equal than $0.15$.}
    \label{fig:comparison}
\end{figure}

Note that $\hat T_{\mathrm{bg}}(\text{smoll})$ and $\hat T_{\mathrm{bg}}(\text{llama})$ are different, since the reference models behave differently depending on the temperatures at which they are tested. This highlights the importance of building  an adequately large set $\mathcal{L}$ in order to obtain a precise estimate of $\overline T_{bg}$ for the system under test. In fact, even if, in general, the trend of the empirical variability distribution computed with different reference systems is similar, its specific values could differ as the different reference LLMs are more or less sensitive to temperature changes, as is clear in Figure \ref{fig:comparison} where the K-S distance between $f_T$ and $\tilde f_{T'}$ is represented, for $T \in \Theta$ and $T' \in \tilde \Theta$.

\subsection{Estimating $T_{\mathrm{bg}}$ for other models}\label{others}
In this section, we present a couple of other experiments aimed at estimating $T_{\mathrm{bg}}$ for other LLMs accessed via external providers (where the environment $I$ is unknown), in analogy with the procedure described in Sections \ref{firstexp} and \ref{extendingL} for \textit{gpt-4.1-nano} provided through Microsoft Azure AI services. In particular we considered the following three other LLMs:
\begin{itemize}
    \item \textit{gemini-2.0-flash} by Google used through the official Gemini API services;
    \item \textit{grok-3-mini} by xAI used through Microsoft Azure AI services;
    \item \textit{claude-sonnet-4} by Anthropic used through Amazon Bedrock services.
\end{itemize}
Each model was run with temperature set to $0$, generating $100$ answers of up to $32$ tokens for the first $30$ prompts in $\Pi$ (see Sections \ref{firstexp} and \ref{extendingL}). As in the case of \textit{gpt-4.1-nano}, we then computed, for every prompt, the maximum fraction of identical answers produced by a given model. Hence we built three discrete distributions (each with $30$ values), denoted $g'$, $g''$, and $g'''$, corresponding respectively to \textit{gemini-2.0-flash}, \textit{grok-3-mini}, and \textit{claude-sonnet-4}, which are the analogue of the distribution $g$ built in Section \ref{firstexp} for \textit{gpt-4.1-nano}. 
As reference variability distributions, we used $f_T'$ and $\tilde f'_{T'}$, obtained by truncating $f_T$ and $\tilde f_{T'}$ to their first $30$ elements (the ones corresponding to the first $30$ prompts in $\Pi$), for $T \in \Theta$ and $T' \in \tilde \Theta$. Thus, by computing the K-S distances of $g'$, $g''$, and $g'''$ with respect to $f_T'$ and $\tilde f'_{T'}$, for $T \in \Theta$ and $T' \in \tilde \Theta$, and then combining them as in Section \ref{extendingL}, we can estimate $T_{\mathrm{bg}}$ for the new models under test. A couple of remarks on these computations:
\begin{itemize}
\item the answers given by \textit{claude-sonnet-4} to each question were identical, making $g'''$ a list of thirty $1$s. In this case, the estimate of $T_{\mathrm{bg}}$ is trivially $0$;
\item the distance between $g''$ and $\tilde f'_{T'}$, for $T' \in \tilde \Theta$, reach its minimum value of $0.167$ for $T' \in \{0.01, 0.02, 0.03\}$. In this case, we set $\hat T_{\mathrm{bg}}(\text{llama}) = 0.02$, i.e. the average of the temperatures minimizing the K-S distance.
\end{itemize}
In Table \ref{estimatesTbg} are summarized the estimates for the tested models, taking in account the remarks above.

\begin{table}[ht]
\centering
\begin{tabular}{lccc}
\toprule
\textbf{Model} & $\hat T_{\mathrm{bg}}(\text{smoll})$ & $\hat T_{\mathrm{bg}}(\text{llama})$ & $\overline T_{\mathrm{bg}}$ \\
\midrule
\textit{grok-3-mini}      & 0.01 & 0.02 & \textbf{0.015} \\
\textit{gemini-2.0-flash} & 0.05 & 0.08 & \textbf{0.065} \\
\textit{gpt-4.1-nano}     & 0.05 & 0.10 & \textbf{0.075} \\
\textit{claude-sonnet-4}  & 0 & 0 & \textbf{0} \\
\bottomrule
\end{tabular}
\caption{Estimates of $T_{\mathrm{bg}}$ as computed in Sections \ref{firstexp}–\ref{extendingL} for \textit{gpt-4.1-nano}, and in Section \ref{others} for the other models.}
\label{estimatesTbg}
\end{table}

\section{Discussion}

The notion of \emph{background temperature} brings several concrete benefits for both evaluation and deployment of LLMs. First, it offers a measurable quantity to capture hidden randomness in inference stacks, converting vague observations of output variability into reproducible metrics; this helps close the gap between what settings are declared “deterministic” (e.g.\ $T=0$) and what is actually observed in practice. Second, it aids diagnostics: by quantifying $T_n(I)$ for different inference-environment parameters (batch size, concurrency, kernel implementation, hardware, or even time of the day and region of deployment), one can identify which aspects of the stack contribute most to instability, and thus target them for engineering improvements. Third, for high-stakes or regulated applications, background temperature enables transparency and trust: one may report that under nominally deterministic settings, the effective randomness is bounded by some small $T_{\mathrm{bg}}$, which supports claims of consistency needed for auditing, compliance, and user trust. Fourth, in research and model evaluation, the measure helps avoid misleading comparisons: if model A slightly outperforms model B under the same settings but with $T_{\mathrm{bg}}$ exceeding that difference, the apparent improvement might simply reflect implementation noise rather than genuine modeling advances. Finally, the concept fosters better engineering practices—motivating adoption of batch‐invariant kernels, deterministic reduction orders, consistent numeric precision and hardware configurations - as suggested by prior work. Overall, background temperature provides a bridge between theory and practice, enabling metrics, engineering tradeoffs, and trustworthiness to advance in parallel.

Issues easiliy anticipated are related to the selection of relevant reference models. While it is likely that they will depend on the specific information domain, their specifications should be subjects to an as much wide as possible consensus in the industry. This may give way to standardisation efforts and certification programmes for models.

%%% Add to your bibliography:

% \bibitem{ouyang2025non}
% Ouyang, Shuyin; Zhang, Jie M.; Harman, Mark; Wang, Meng (2025). “An Empirical Study of the Non-Determinism of ChatGPT in Code Generation.” ACM Transactions on Software Engineering and Methodology. :contentReference[oaicite:0]{index=0}

% \bibitem{tmlab2025defeating}
% Thinking Machines Lab (2025). “Defeating Nondeterminism in LLM Inference.” Blog. :contentReference[oaicite:1]{index=1}

\section{Conclusion}

In this note we introduced \emph{background temperature} as a concise lens on hidden randomness in LLM inference. The notion of background temperature reframes a practical pain point - apparent nondeterminism at $T=0$—in terms familiar to modeling and evaluation. By \emph{measuring} $T_{\mathrm{bg}}$, practitioners can make informed choices about evaluation protocols, safety margins, and infrastructure investments. Reporting $T_{\mathrm{bg}}$ (and its higher order statistics) alongside accuracy/throughput could become a valuable practice for transparent model documentation.
The definition and operationalisation we gave of $T_{\mathrm{bg}}$ captures implementation-induced variability when nominal $T=0$. We outline a measurement protocol and a set summarized engineering strategies to reduce $T_{\mathrm{bg}}$ as well as a set of small pilot experiments to illustrate the potential of the concept. Future work should include standardizing metrics, open datasets of prompts sensitive to implementation noise, and community benchmarks for deterministic inference.

\bibliographystyle{plain}
\bibliography{biblio}
\end{document}